%% file: acl_latex.tex
\documentclass[11pt]{article}

\usepackage[final]{acl}
\usepackage{times}
\usepackage{latexsym}
\usepackage[T1]{fontenc}
\usepackage[utf8]{inputenc}
\usepackage{microtype}
\usepackage{inconsolata}
\usepackage{graphicx}
\usepackage{makecell}

\usepackage{multirow}
\usepackage{booktabs}
\usepackage{amsmath,amsfonts}
\usepackage{algorithmic}
\usepackage{algorithm}
\usepackage{bm}
\usepackage{subfigure}
\usepackage{float}

\usepackage[table]{xcolor}

\usepackage{amssymb}

\title{$\bm{\mathcal{R}}^3$: Advertisement Compliance $\bm{\mathcal{R}}$\!ectification via Group-$\bm{\mathcal{R}}$\!elative Experience Extractor and Curriculum $\bm{\mathcal{R}}$\!einforcement}

\author{
Yuan Chen\thanks{Equal Contribution.} ~~Zhenyu Hu\footnotemark[1] ~~Mengge Xue\thanks{Project Leader.} ~~Te Cao ~~{\bf Liqun Liu}\thanks{Corresponding Author.} \\ ~~{\bf Peng Shu} ~~{\bf Huan Yu} ~~{\bf Jie Jiang} \\
Tencent \\
\texttt{\{izayoiychen, mapleshu, berryxue, rosaliecao, liqunliu\}@tencent.com} \\ \texttt{\{archershu, huanyu, zeus\}@tencent.com}
}

\begin{document}
\maketitle

\input{body/abstract}

\input{body/introduction}

\begin{figure*}[t]
\centering
  \includegraphics[width=1.0\textwidth, trim=0 15mm 5mm 0]{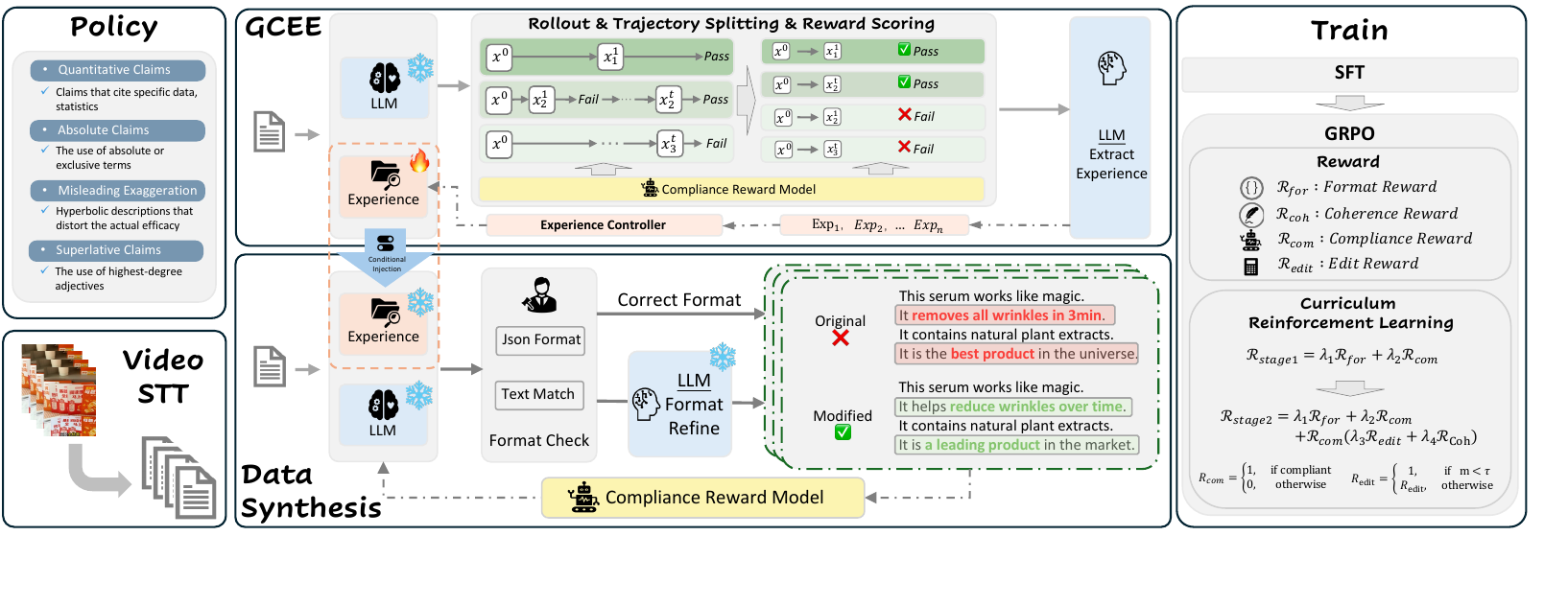} 
  \caption {\textbf{Overview of $\bm{\mathcal{R}^3}$.} Taking non-compliant video ads and violation policies as input, the \textbf{Experience-driven Data Synthesis} employs the \textbf{Group-Relative Compliance Experience Extractor (GCEE)} to extract compliance experience from rectification trajectories for high-quality supervision. The model is initialized via supervised fine-tuning and further optimized using a \textbf{Curriculum Reinforcement Learning} strategy with hierarchical rewards.}
  \label{fig:overview}
\end{figure*}

\input{body/Related_Work}

\input{body/method}

\input{body/deployment}

\input{resources/tables/table1}

\input{body/experiments}
\input{body/conclusion}

\section{Limitations}

Our system is optimized against a specific moderation system and a fixed set of policy guidelines used during training, which may reduce flexibility when the moderation system, decision boundary, or rule set changes over time. While our experience-driven data synthesis and  curriculum reinforcement learning improve robustness within the covered policy scope, adapting to newly introduced or rapidly evolving rules may still require additional data regeneration and re-alignment. As future work, we plan to explore continual and modular policy alignment strategies that can rapidly incorporate rule updates with minimal re-training.

\section{Ethical Statement}
This research was conducted in strict adherence to ethical guidelines and data privacy regulations. The industrial dataset utilized in this study was derived from a production environment and was rigorously desensitized to ensure the anonymity of advertisers and users. The non-compliant advertisement examples presented herein are strictly for illustrative scientific analysis and do not reflect the views or values of the authors or the affiliated platform. All resources and methodologies are intended solely for academic research.

\bibliography{custom}

\appendix

\input{body/7_appendix}

\end{document}

%% file: body/abstract.tex
\begin{abstract}

Rigorous content moderation is crucial for online advertising but leads to millions of daily rejections. This scale renders manual rectification infeasible, particularly for video advertisements.
However, existing safety-driven methods often suffer from aggressive over-editing, which compromises the advertiser's original semantic intent merely to satisfy compliance.
In this work, we target the rectification of textual violations in video ads, covering both speech transcripts and on-screen text. We propose $\bm{\mathcal{R}}^3$, a novel framework designed to harmonize compliance with original semantic intent preservation.
Our approach integrates three key innovations: (1) an experience-driven data synthesis framework that bootstraps high-quality supervision via group-\textbf{R}elative compliance experience extractor; 
(2) a curriculum \textbf{R}einforcement learning strategy with hierarchical rewards designed to enforce compliance while maximizing semantic consistency;
and 
(3) a comprehensive video \textbf{R}ectification framework seamlessly integrating text recognition, rewriting, and re-rendering for industrial deployment. Extensive experiments on industrial datasets and online A/B testing demonstrate that $\mathcal{R}^3$ significantly outperforms state-of-the-art baselines, achieving an optimal trade-off between violation rectification and intent preservation.

\end{abstract}

%% file: body/introduction.tex
\section{Introduction}

Advertising serves as a cornerstone of the digital economy, acting as the primary engine for revenue and growth across online platforms~\cite{rathee2024sustainability,campbell2025diversity}. 
In pursuit of strict regulatory compliance and user safety, these platforms impose rigorous content moderation policies ~\cite{ji-etal-2025-raven,ji-etal-2025-raven-pinpointing,madio2025content}. 
However, advertisers often struggle to navigate the complexity of these moderation rules, resulting in millions of advertisements being rejected daily. 
Consequently, it is imperative for online platforms to assist advertisers in automatically rectifying ad material, thereby unlocking ad supply and enhancing the overall advertiser experience~\citep{xia-etal-2025-reimagining}. With the advent of the 5G era, video has emerged as the predominant medium for information consumption, a trend particularly evident in online advertising. 
Consequently, violations within video advertisements constitute a significant proportion of overall content compliance issues. 
While these violations span both visual and textual modalities, this work focus specifically on the rectification of textual violations within video ads.

While recent advancements in Large Language Models (LLMs)~\citep{touvron2023llama,qwen3,openai2023gpt4} have bolstered capabilities in content moderation and compliance rectification~\cite{pi2024mllmprotector,laugier-etal-2021-civil}, directly applying them to video ad rectification remains non-trivial. 
Specifically, deploying general-purpose or naively fine-tuned models for this task faces significant challenges: 
1) \textbf{Inadequate compliance}: Ad moderation policies are voluminous and highly context-dependent, thus that general-purpose models often fail to grasp the nuanced boundaries of moderation rules via direct prompting, leading to frequent hallucinations or missed detections of subtle violations. This necessitates domain-specific alignment, yet standard Supervised Fine-Tuning (SFT) is hindered by data availability; 
2) \textbf{Prohibitive Annotation Costs}: The prerequisite data annotation phase for supervised fine-tuning is severely constrained by the complexity and rapid evolution of ad content moderation rules. The substantial cognitive burden involved in comprehending these policies, coupled with the resulting low annotation consistency, renders large-scale manual annotation practically infeasible~\citep{ji-etal-2025-raven}. 
3) \textbf{Compromised semantic intent}: violation content rewriting is inherently a multi-objective optimization task that demands balancing compliance, semantic intent preservation, and coherence. Naively fine-tuned models, however, tend to over-optimize for compliance, leading to excessive alterations that deviate from the original intent and diminish the advertising effectiveness.

To address the aforementioned challenges, we introduce ${\mathcal{R}}^3$, a comprehensive video rectification framework designed to rectify textual violations in video advertisements, covering both speech transcripts and on-screen text. By harmonizing strict compliance with the preservation of the advertiser's original semantic intent, ${\mathcal{R}}^3$ automates the full lifecycle of video rectification—from multi-modal content extraction to final re-rendering. Our main contributions are summarized as follows:

1) Experience-driven Data Synthesis: We propose an experience-driven synthesis framework that bootstraps high-quality supervision from advanced LLMs. By introducing group-relative compliance experience extractor, it effectively addresses data scarcity and covers complex edge cases where naive prompting typically fails.

2) Curriculum RL with Hierarchical Rewards: The task of violation textual content rectification demands the simultaneous satisfaction of multiple, often competing objectives: compliance, semantic intent preservation, and coherence. Optimizing these concurrently poses significant challenges for standard Reinforcement Learning (RL) paradigms. We address this by introducing a curriculum~\citep{bengio2009curriculum,ko2022curriculum} reinforcement learning framework with hierarchical rewards. This strategy progressively optimizes the model to navigate the intrinsic conflict between enforcing strict compliance and maintaining semantic intent, avoiding the over-editing in existing methods.

3) Holistic Video Rectification Framework: We present a non-compliant video ads rectification system validated within a live online advertising platform. Going beyond violation text rewriting, we engineer a pipeline that seamlessly integrates text recognition, rewriting, and re-rendering for industrial-scale deployment. This enables the automated rectification of non-compliant videos while preserving their original audio-visual fidelity.

We evaluated ${\mathcal{R}}^3$ using industry datasets and verified its effectiveness on online advertisement platform. 
Comparative analysis reveals that our system significantly outperforms general-purpose models-ranging from the open-source Qwen3~\cite{qwen3} to the advanced Gemini3-Flash~\cite{google2025gemini3}. 
Crucially, ${\mathcal{R}}^3$ demonstrates a superior performance to balance compliance with semantic intent preservation, resulting in a significantly improved final successful rectification rate.

%% file: body/Related_Work.tex
\section{Related Work}

\subsection{Alignment for Large Language Models}
RL has become the standard for LLM alignment~\citep{christiano2017deep}, with recent advances like Group Relative Policy Optimization (GRPO)~\citep{shao2024deepseekmath} improving efficiency by estimating advantages from response groups rather than separate value functions. Beyond general alignment, RL has shown promise in complex decision-making tasks such as content moderation~\citep{ji-etal-2025-raven-pinpointing,ji-etal-2025-raven}. Despite these successes, existing methods~\citep{wang2021rl4rs,cai2023constrained} primarily focus on open-ended or discriminative tasks. In our work, we apply RL to video rectification, which bridges this gap by designing a sophisticated, curriculum reward mechanism.

\subsection{Generative AI and Content Moderation in Online Advertising}
Generation works leverage generative models to maximize commercial metrics~\cite{deng2025onerec,zhang2025gpr,wu2024survey}, often creating content from scratch rather than preserving existing intent. Meanwhile, ~\citet{ji-etal-2025-raven-pinpointing} establish robust baselines for risk detection and localization but stop short of correction. This creates a disjointed industrial pipeline where rejected ads lack automated rectification. Our work bridges this gap. Distinct from over-editing methods like ~\citep{zhang2025safeeditor}, we employ curriculum RL to rewrite violative content, achieving an optimal trade-off between compliance and intent preservation.

\subsection{Training-Free Methods}
Existing inference-time methods, including in-context learning (ICL)\citep{brown2020language}, iterative refinement\citep{madaan2023self,shinn2023reflexion}, and feedback-driven frameworks~\citep{song2025reward,yuksekgonul2025optimizing,monea2024llms}, typically target within-sample improvements. Compared to the hierarchical, off-policy design of Agent KB~\citep{tang2025agent}, Training-Free GRPO~\citep{training_free_grpo} adopts a global perspective, refining a shared experience library via multi-epoch learning akin to traditional RL. Inspired by Training-Free GRPO, our approach performs task-aware trajectory split and extracts experience by contrasting successful and failed trajectories.

%% file: body/method.tex
\section{Method}

\subsection{Problem Formulation}

We formulate violation textual content rectification as a constrained rewriting task defined by the tuple $\mathcal{I}=(x, \mathcal{G})$. Here, $x$ represents the non-compliant video textual content, $\mathcal{G}$ denotes the violation policies provided by the online violation detection model. 
We segment $x$ into sentence units $\mathcal{S}(x)=[u_1, u_2, \dots, u_n ]$ using standard punctuation-based splitting and train a policy $\pi_\theta(y|x, \mathcal{G})$ to generate a structured edit list $y=\{(u_k, v_k) \mid u_k \in \mathcal{N}\}$, where $\mathcal{N} \subset \mathcal{S}(x)$ represents the subset of sentences identified as violations.Specifically, each pair in $y$ indicates that a non-compliant sentence $u_k$ is substituted by its compliant revision $v_k$. The final rectified text $x^\prime$ is obtained via a deterministic mapping $\mathcal{A}(x, y)$, which applies these substitutions to the original sequence.
Our objective is to produce a rectified output $x^\prime$ that ensures compliance, while preserving the semantic intent of $x$ by optimizing for linguistic coherence and minimizing edits.

\subsection{Experience-driven Data Synthesis}
\label{sec:data_generation}
Manual rectification of non-compliant ad material is labor-intensive, time-consuming, and unscalable.
While Advanced LLMs possess strong constrained rewriting capabilities, they often exhibit inadequate compliance, lacking the nuanced understanding of specific moderation policies.
To bridge this gap, we propose an automated, experience-driven data synthesis framework that bootstraps high-quality supervision. 
Specifically, tractable samples are efficiently processed by an advanced LLM Gemini3-Flash, while for intractable samples where direct prompting fails, we incorporate a Group-relative Compliance Experience Extractor.

\paragraph{Group-relative Compliance Experience Extractor.} This module is an in-context reinforcement learning paradigm inspired by Training-Free GRPO~\citep{training_free_grpo}. Adopting the core GRPO mechanism, it performs rollouts and computes group advantages, yet distinguishes itself by learning through natural language feedback rather than parameter updates. Operating within an iterative refinement framework (starting from $x^{0}=x$ to generate $x^{t+1} = \mathcal{A}(x^{t}, y^{t})$), the extractor pinpoints the pivotal violation-to-compliance transition—where a persistent non-compliant $x^{t}$ becomes compliant $x^{t+1}$. It constructs a high-advantage contrastive pair by comparing the failed trajectory $x^{0} \to x^{t}$ with the successful trajectory $x^{0} \to x^{t+1}$, prompting the LLM to explicitly articulate the semantic rationale behind the correction. This experience is stored in a dynamic buffer, where an LLM-based controller compares newly extracted experiences with existing entries and decides whether to keep, revise, or discard them to reduce redundancy and conflicts. The refined experience are then injected into subsequent generations. An example can be found in Appendix~\ref{sec:app_example_GCEE}.

\paragraph{Conditional Experience Injection.} However, the extracted experience inherently prioritizes compliance, often biasing the model toward aggressive rewriting that deviates from the original semantic intent. To resolve this trade-off, as shown in Figure~\ref{fig:overview}, we implement a conditional injection mechanism. We first employ a direct prompting strategy using Gemini3-Flash, activating experience injection only if the initial attempt fails. This strategy prioritizes minimal edits on tractable instances and reserves experience injection for intractable ones, thereby producing a high-coverage training corpus that strictly adheres to multi-objectives.

\subsection{Training}

We train ${\mathcal{R}}^3$ on top of Qwen3-8B by integrating SFT with GRPO. Specifically, the SFT stage serves to instill moderation policies and align the model with the required structured output format. Subsequently, GRPO further optimizes for compliance and semantic intent preservation using curriculum RL with hierarchical rewards.

\subsubsection{Supervised Fine-Tuning}
SFT enforce the edit-list schema and condition rectifications. This stage minimizes format violations and injects foundational compliance knowledge, establishing a stable initial state to facilitate efficient exploration in the subsequent RL stage.

\subsubsection{Rewards Design}
The reward function R optimizes the model’s RL performance across four dimensions.

\paragraph{Format Reward ($\bm{R_{for}}$).} This term enforces structural executability. Beyond JSON parsability, every $u_k$ selected for modification must exactly match the source text (i.e., $u_k\in\mathcal{S}(x)$).

\paragraph{Compliance Reward ($\bm{R_{com}}$).}
We measure compliance using the online violation detection model. The reward is binary: $R_{com} \in\{0,1\}$. This term acts as the primary driver for satisfying compliance.

\paragraph{Minimal-edit Reward ($\bm{R_{edit}}$).}
We quantify modification magnitude by the count of edited units $m=|y|$ relative to $n=|\mathcal{S}(x)|$. Minimizing $m$ is crucial for: (1) Advertisers prioritize retaining the original narrative structure and minimizing interventions; (2) Limiting text changes reduces duration mismatches in downstream Text-To-Speech (TTS), preventing synchronization failures. However, an unbounded penalty on $m$ creates an optimization imbalance: over-optimizing minimalism on tractable samples penalizes the exploration required for intractable ones. Thus, we introduce a tolerance threshold $\tau$ as an edit margin to establish a penalty-free zone for essential modifications:
\begin{equation}
\label{eq:reward_edit_tilde}
    R_{\text{edit}}=
    \begin{cases}
    1, & m \le \tau,\\
    1-\frac{m-\tau}{n-\tau}, & m > \tau.
    \end{cases}
\end{equation}

\paragraph{Coherence Reward (\bm{$R_{coh}$})}
To ensure linguistic fluency and semantic preservation, we employ an LLM-as-a-judge to provide dense feedback. 

For each pair $(u_k, v_k)$, the judge assigns a scalar score $\phi(x, u_k, v_k) \in \{0, 1\}$.
The final sentence-level reward is the average of these scores:
\begin{equation}
\label{eq:reward_coh_tilde}
    R_{\text{coh}}=\frac{1}{m}\sum_{k=1}^{m}\phi(x, u_k, v_k).
\end{equation}

The overall reward $R$ is:
\begin{equation}
    \label{eq:all_reward}
    R = \lambda_1 R_{for} + \lambda_2 R_{com} + R_{com} \cdot (\lambda_3 R_{edit} + \lambda_4 R_{coh}),
\end{equation}

where $\lambda_1,\lambda_2,\lambda_3, \lambda_4$ are stage-dependent coefficients.Specifically, in Stage 1 we set $(\lambda_1, \lambda_2, \lambda_3, \lambda_4) = (0.3, 5.0, 0, 0)$, and in Stage 2 we use $(0.1, 5, 0.4, 0.5)$. To prevent reward hacking, we gate $R_{edit}$ and $R_{coh}$ using the compliance reward $R_{com}$. This ensures that auxiliary objectives are only pursued once the primary compliance goal is met.

\subsubsection{Curriculum RL with Hierarchical Rewards}
\label{sec:stages}

While SFT provides a strong initialization, it relies on static supervision, which struggles to balance the often conflicting objectives of strict compliance and semantic preservation. To address this, we employ RL with a hierarchical reward system. 
However, directly optimizing for these competing goals remains challenging: if edit penalties are introduced too early, the policy tends to collapse into a local optimum of minimal edits to avoid punishment, failing to explore necessary but extensive rewrites. To overcome this exploration hurdle, we design a two-stage curriculum:

\begin{figure*}[t]
\centering
  \includegraphics[width=1.0\linewidth]{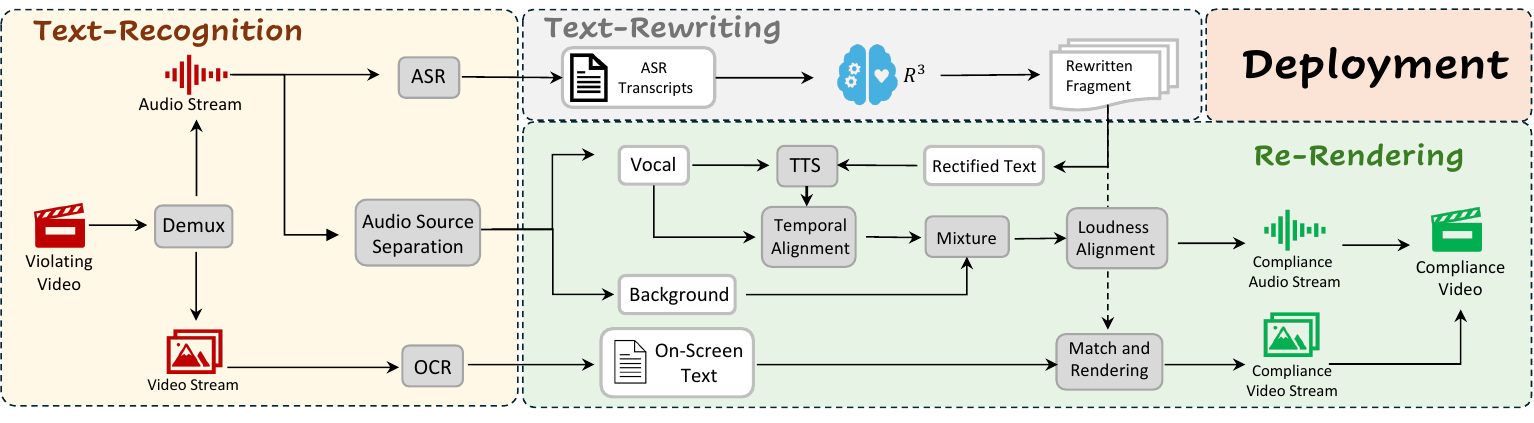} 
  \caption {The deployment workflow of $\mathcal{R}^3$. Illustrating the automated pipeline for rectification}
  \label{fig:deploy_pipeline}
\end{figure*}

\paragraph{Stage 1: Compliance Alignment.}

In the initial stage, we exclusively prioritize compliance to encourage bold exploration. We set the coefficients $\lambda_3$ and $\lambda_4$ to zero while assigning a high value to $\lambda_2$. This configuration compels the policy model to focus solely on satisfying compliance, incentivizing extensive rewriting. While this may temporarily lead to over-editing, it establishes a critical high-recall baseline, ensuring the model learns the necessary editing to achieve high compliance rate.

\paragraph{Stage 2: Quality Refinement.}
Once the model achieves a stable compliance rate, we transition to stage 2 by activating the soft objectives ($\lambda_3, \lambda_4 > 0$) while maintaining $\lambda_2$ as the dominant term. This stage acts as a regularization step: within the manifold of compliant solutions, the policy model is guided to select those that minimize text alterations while preserving coherence. This effectively refines the aggressive editing behavior learned in stage 1, converging toward the Pareto frontier of compliance and semantic intent preservation.

%% file: body/deployment.tex
\section{Deployment}
\label{sec:deployment}

To bridge the gap between algorithmic rewriting and industrial production, we integrate $\mathcal{R}^3$ into an automated video rectification workflow (Figure~\ref{fig:deploy_pipeline}). The pipeline comprises three phases: (1) Text-Recognition, which demultiplexes the video, isolates vocals via audio source separation, and extracts both speech transcripts and on-screen text; (2) Text-Rewriting, where $\mathcal{R}^3$ generates a structured edit list to ensure compliance; and (3) Re-Rendering, which reconstructs both auditory and visual content.

The final Re-Rendering stage is critical for preserving the quality of the rectified video.We employ zero-shot voice cloning to preserve the speaker's timbre. A key challenge is temporal alignment between synthesized speech and the original video timeline. We therefore first adjust silent intervals and, when necessary, apply time-stretching to the generated speech so that the updated segment matches the target duration. For loudness alignment, we employ an iterative dynamic range control procedure: we first estimate the loudness gap between the synthesized speech and the original vocal track, then apply gain adjustment with peak limiting, and iteratively compensate for the attenuation introduced by the limiter until the mixed audio reaches the target loudness. Finally, subtitles are re-rendering using OCR-derived coordinates and the updated audio is merged back into the video, producing a compliant video that preserves the original production layout and audio-visual fidelity.

%% file: resources/tables/table1.tex
\begin{table*}[h]

    \centering
    \resizebox{0.85\textwidth}{!}{%
    \begin{tabular}{lccccccc >{\columncolor{lightgray!50}} c}
        \toprule

        \multirow{2}{*}[-0.3em]{\textbf{Model}} & 
        \multirow{2}{*}[-0.3em]{\makecell{\textbf{Qua.C} \\ \textbf{ComR$\uparrow$}}} & 
        \multirow{2}{*}[-0.3em]{\makecell{\textbf{Abs.C} \\ \textbf{ComR$\uparrow$}}} & 
        \multirow{2}{*}[-0.3em]{\makecell{\textbf{Mis.E} \\ \textbf{ComR$\uparrow$}}} & 
        \multirow{2}{*}[-0.3em]{\makecell{\textbf{Sup.C} \\ \textbf{ComR$\uparrow$}}} &
        \multicolumn{4}{c}{\textbf{Average}} \\
        \cmidrule(lr){6-9}

         & & & & & \textbf{ComR$\uparrow$} & \textbf{AvgE$\downarrow$} & \textbf{CohR$\uparrow$} & \textbf{QRR$\uparrow$} \\
        \midrule
        Qwen3-8B & 69.32\% & 67.95\% & 31.10\% & 70.88\% & 50.28\% & 8.2 & 45.07\% & 22.66\% \\
        Gemini3-Flash & 83.67\% & 86.32\% & 65.99\% & 96.70\% & 77.05\% & \textbf{6.24} & \textbf{97.98\%}  & 74.43\% \\
        Gemini3-Flash with GCEE & 90.04\% & 90.17\% & 75.58\% & 98.35\% & 84.84\% & 8.63 & 89.41\% & 75.21\% \\
        Qwen3-8B-SFT & 91.63\% & 94.87\% & 69.77\% & 98.90\% & 82.58\% & 7.7 & 87.48\% & 72.23\% \\
        \midrule
        $\bm{\mathcal{R}^3}$ & \textbf{93.63\%} & \textbf{96.58\%} & \textbf{76.16\%} & \textbf{98.90\%} & \textbf{85.50\%} & 7.49 & 94.70\% & \textbf{81.02\%} \\
        \bottomrule
    \end{tabular}%
    }
    \caption{\textbf{Performance comparison.} We report ComR on four violation policies. Additionally, we report the average performance on four metrics: ComR, AvgE, CohR, and QRR. $\mathcal{R}^3$ achieves the best performance.}
  \label{tab:model_performance_comparison}
\end{table*}

%% file: body/experiments.tex
\section{Experiments}

\subsection{Experimental Setup}

\subsubsection{Datasets}
To assess the efficacy of $\mathcal{R}^3$ within a realistic industrial setting, we constructed a proprietary dataset derived from a production moderation pipeline. The dataset is partitioned into a training set of 11k samples and a held-out test set of 1k samples. To ensure a rigorous evaluation of robustness across challenging compliance scenarios, we focus on four major violation policies: \textbf{Quantitative Claims (Qua.C)}, \textbf{Absolute Claims (Abs.C)}, \textbf{Misleading Exaggeration (Mis.E)}, and \textbf{Superlative Claims (Sup.C)}, detailed definitions and examples are provided in the Appendix \ref{sec:app_definition}.

\subsubsection{Metrics}
To evaluate the performance of our method, we employ four metrics: (1) \textbf{Compliance Rate (ComR)}: The percentage of rectified samples which are compliant. (2) \textbf{Average Edit (AvgE)}: The average number of modified sentence units per example. (3) \textbf{Coherence Rate (CohR)}: The proportion of samples deemed linguistically fluent and semantically consistent. We employ Gemini3-Flash as an evaluator, instructing it to compare the rectified content against the original text. (4) \textbf{Qualified Rectification Rate (QRR)}: The fraction of samples that are both compliant and coherent. In practice, advertiser tolerance for text alterations is highly subjective, making a universal threshold for acceptable edits impractical.
Thus, QRR indicates whether a rectification is fundamentally usable, while AvgE independently measures the intervention cost required to achieve this qualification.

\input{resources/tables/table_abtest}

\input{resources/tables/table_ablation}

\subsection{Offline Testing}
Table~\ref{tab:model_performance_comparison} presents a comprehensive evaluation of $\mathcal{R}^3$ against baselines, including the Qwen3-8B, Gemini3-Flash (with and without GCEE), and Qwen3-8B-SFT. $\mathcal{R}^3$ achieves state-of-the-art performance with the highest QRR and remarkably competitive AvgE. Specifically, compared to the Qwen3-8B-SFT, $\mathcal{R}^3$ yields a consistent improvement in ComR (+2.92\%) and a substantial margin in QRR (+8.79\%). This confirms that integrating curriculum GRPO effectively steers the policy toward the optimal Pareto frontier of strict compliance and semantic intent preservation. Furthermore, $\mathcal{R}^3$ ranks first across all four fine-grained violation policies. While Gemini3-Flash exhibits highly conservative editing behavior, $\mathcal{R}^3$ significantly outperforms it in overall quality, surpassing this baseline by +8.45\% in ComR and +6.59\% in QRR. Although augmenting Gemini3-Flash with GCEE boosts its compliance to 84.84\%, it noticeably increases AvgE to 8.63 and limits the overall qualification. $\mathcal{R}^3$ maintains a definitive lead, validating the superiority of a dedicated rectification policy over directly prompting LLMs.

\subsection{Online A/B Testing}
To assess the practical viability of our approach in a production environment, we conducted a 3-day online A/B test on a live advertisement platform, benchmarking $\mathcal{R}^3$ against Qwen3-8B-SFT. We report adoption rates normalized against the baseline, as absolute figures are commercially sensitive. $\mathcal{R}^3$ yields substantial improvements (Table~\ref{tab:online_sample_average}), with a +2.62\% ComR gain  and a relative adoption increase of +21\%. Given that both models exhibit comparable AvgE, the boost in adoption aligns directly with $\mathcal{R}^3$'s higher QRR. This demonstrates that evaluating QRR in conjunction with AvgE effectively predicts real-world advertiser acceptance.

\begin{figure}[h]
\centering
  \includegraphics[width=0.83\linewidth]{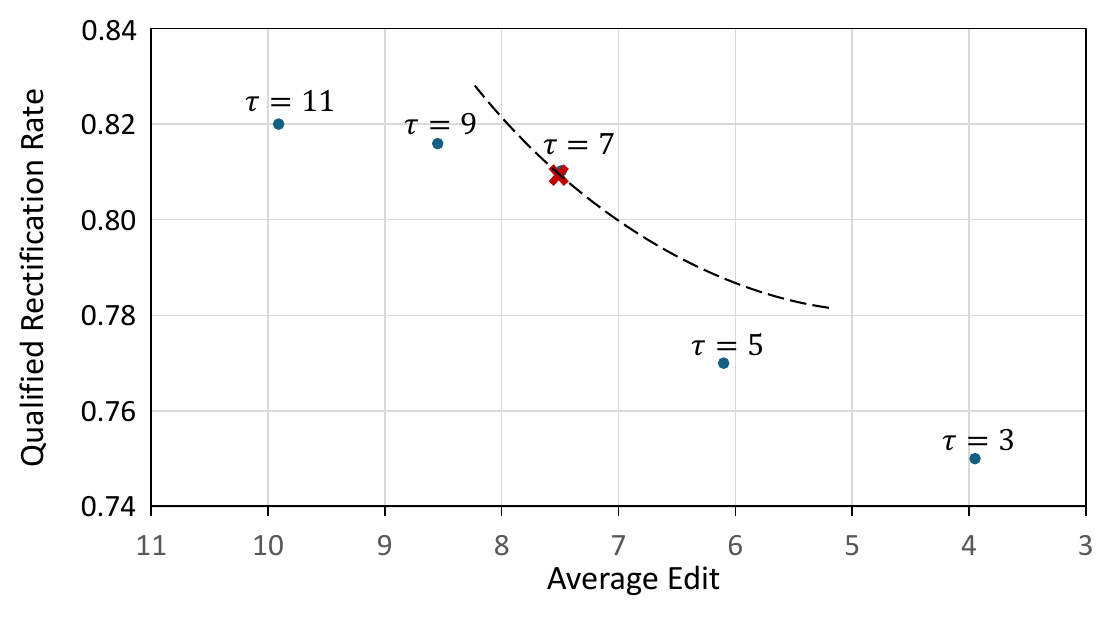} 
  \caption {Impact of the tolerance threshold $\tau$.}
  \label{fig:tolerance}
\end{figure}

\subsection{Ablation Study}

\subsubsection{Study on Rewards and Curriculum RL}
We evaluate the contributions of distinct reward components and the curriculum strategy in Table~\ref{tab:ablation_study}. Regarding reward design, relying solely on the Compliance Reward guarantees compliance but induces excessive alterations. Integrating the Minimal-edit Reward effectively constrains this over-editing behavior, while the Coherence Reward is indispensable for preserving linguistic fluency and semantic intent. Regarding the training strategy, comparing single-stage optimization against our two-stage curriculum confirms that the latter yields a superior trade-off.

\subsubsection{Study on the Tolerance Threshold}
Figure~\ref{fig:tolerance} depicts the impact of the tolerance threshold $\tau$ . We observe that a low $\tau$ overly restricts edits, compromising compliance, whereas an excessively high $\tau$ improves compliance but suffers from over-editing. This dynamic constitutes a Pareto optimality problem. As shown in Figure~\ref{fig:tolerance}, setting $\tau=7$ yields the most favorable equilibrium between rewriting intensity and compliance quality.

%% file: resources/tables/table_abtest.tex
\begin{table}[h]
    \centering
    \resizebox{0.7\columnwidth}{!}{%
    \begin{tabular}{lcc}
        \toprule
        \multirow{2}{*}[-0.3em]{\textbf{Model}} & \multicolumn{2}{c}{\textbf{Online Sample Average}} \\
        \cmidrule(lr){2-3}
         & \textbf{ComR$\uparrow$} & \textbf{AR$\uparrow$} \\
        \midrule
        Qwen3-8B-SFT & 83.91\% & 1.0 \\
        $\bm{\mathcal{R}^3}$    & 86.53\% & 1.21 \\
        \bottomrule
    \end{tabular}%
    }

     \caption{Performance comparison on online A/B test. AR denotes the Advertiser Adoption Rate}
    \label{tab:online_sample_average}
\end{table}

%% file: resources/tables/table_ablation.tex
\begin{table}[!ht]
\centering

\newcommand{\cmark}{\checkmark}
\resizebox{1\linewidth}{!}{
\begin{tabular}{ccccc|cccc}
\toprule

\multirow{2}{*}[-0.3em]{\makecell{\textbf{SFT}}} & 
\multirow{2}{*}[-0.3em]{\makecell{\textbf{$\bm{R_{com}}$}}} & 
\multirow{2}{*}[-0.3em]{\makecell{\textbf{$\bm{R_{edit}}$}}} & 
\multirow{2}{*}[-0.3em]{\makecell{\textbf{$\bm{R_{coh}}$}}} & 
\multirow{2}{*}[-0.3em]{\makecell{\textbf{Curr.} \\ \textbf{RL}}} &
\multicolumn{4}{c}{\textbf{Average}} \\
\cmidrule(lr){6-9}
 & & & & & \textbf{ComR$\uparrow$} & \textbf{AvgE$\downarrow$}&\textbf{CohR$\uparrow$} & \textbf{QRR$\uparrow$} \\

\midrule

\cmark & - & - & - & - & 82.58\% & 7.70 & 87.48\% & 72.23\% \\
\cmark & \cmark & - & - & - & \textbf{89.94\%} & 11.28 & 86.14\% & 77.47\% \\
\cmark & \cmark & \cmark & - & - & 83.43\% & \textbf{6.31} & 87.95\% & 73.37\% \\
\cmark & \cmark & \cmark & \cmark & - & 83.14\% & 6.56 & 92.84\% & 77.19\% \\
\cmark & \cmark & \cmark & \cmark & \cmark & 85.50\% & 7.49 & \textbf{94.70\%} & \textbf{81.02\%} \\
\bottomrule
\end{tabular}
}
\caption{Ablation study on different reward components and curriculum learning (Curr. RL) strategies.}

\label{tab:ablation_study}
\end{table}

%% file: body/conclusion.tex
\section{Conclusion}
We present $\mathcal{R}^3$, a comprehensive video textual rectification framework designed to harmonize strict compliance with semantic intent preservation. By employing experience-driven data synthesis and curriculum reinforcement learning, $\mathcal{R}^3$ achieves state-of-the-art gains in both compliance rate and qualified rectification rate on industrial benchmarks. Furthermore, online A/B testing validates its practical viability and superior advertiser adoption in real-world production environments.

%% file: body/7_appendix.tex
\section{Dataset Statistics and Violation Policy Definitions}
\label{sec:appendix}
\subsection{Data Distribution}

\begin{figure}[h]
    \centering
    \subfigure[Training Set Distribution]{
        \label{fig:train_dist}
        \includegraphics[width=0.8\linewidth]{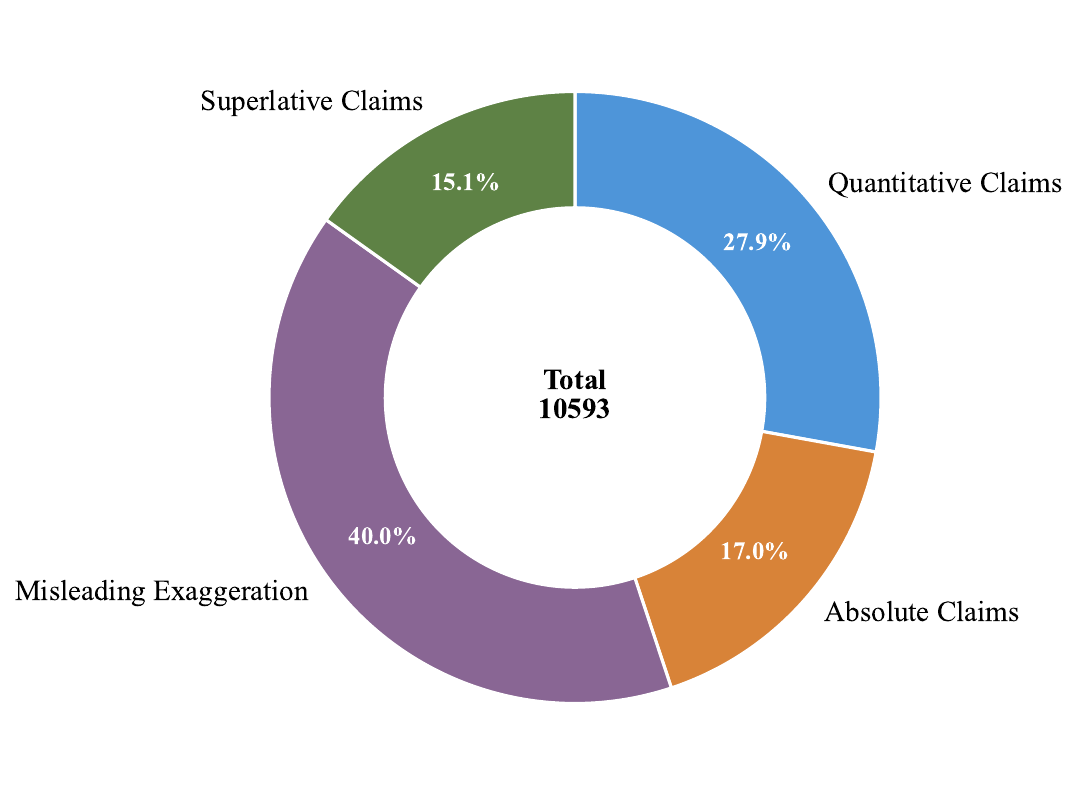}
    }
    \subfigure[Test Set Distribution]{
        \label{fig:test_dist}
        \includegraphics[width=0.8\linewidth]{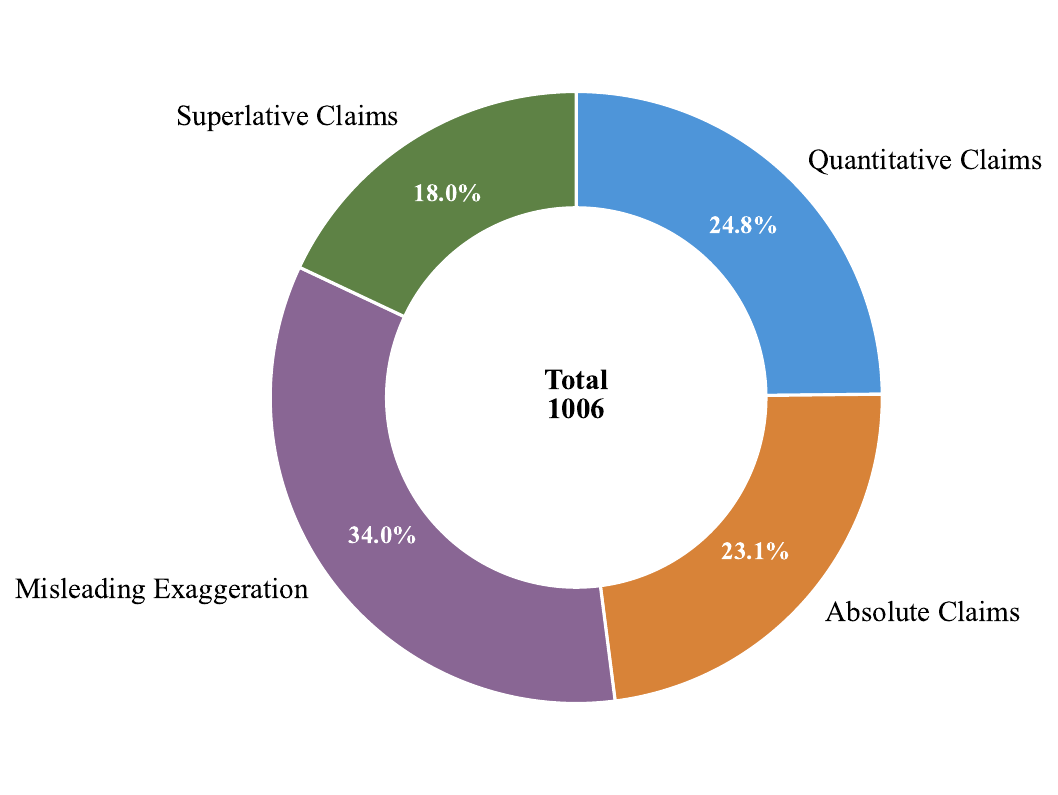} 
    }
    
    \caption{Distribution of violation policies in the training and test datasets.}
    \label{fig:data_dist}
\end{figure}

Figure \ref{fig:data_dist} illustrates the distribution of the violation policies across our industrial dataset.

\subsection{Violation Policy Definitions}
\label{sec:app_definition}
We align our violation policies with the production moderation standards of the online platform. Table \ref{tab:violation_def} presents desensitized definitions and illustrative examples, intended solely to capture the general linguistic characteristics of each violation policy.
\input{resources/tables/table_supp_violation_example}

\section{Details of Experience-Driven Data Synthesis}

\subsection{Prompt for Standard Rectification (SFT \& Data Synthesis)}
\begin{figure*}[t]
  
  \includegraphics[width=1\linewidth]{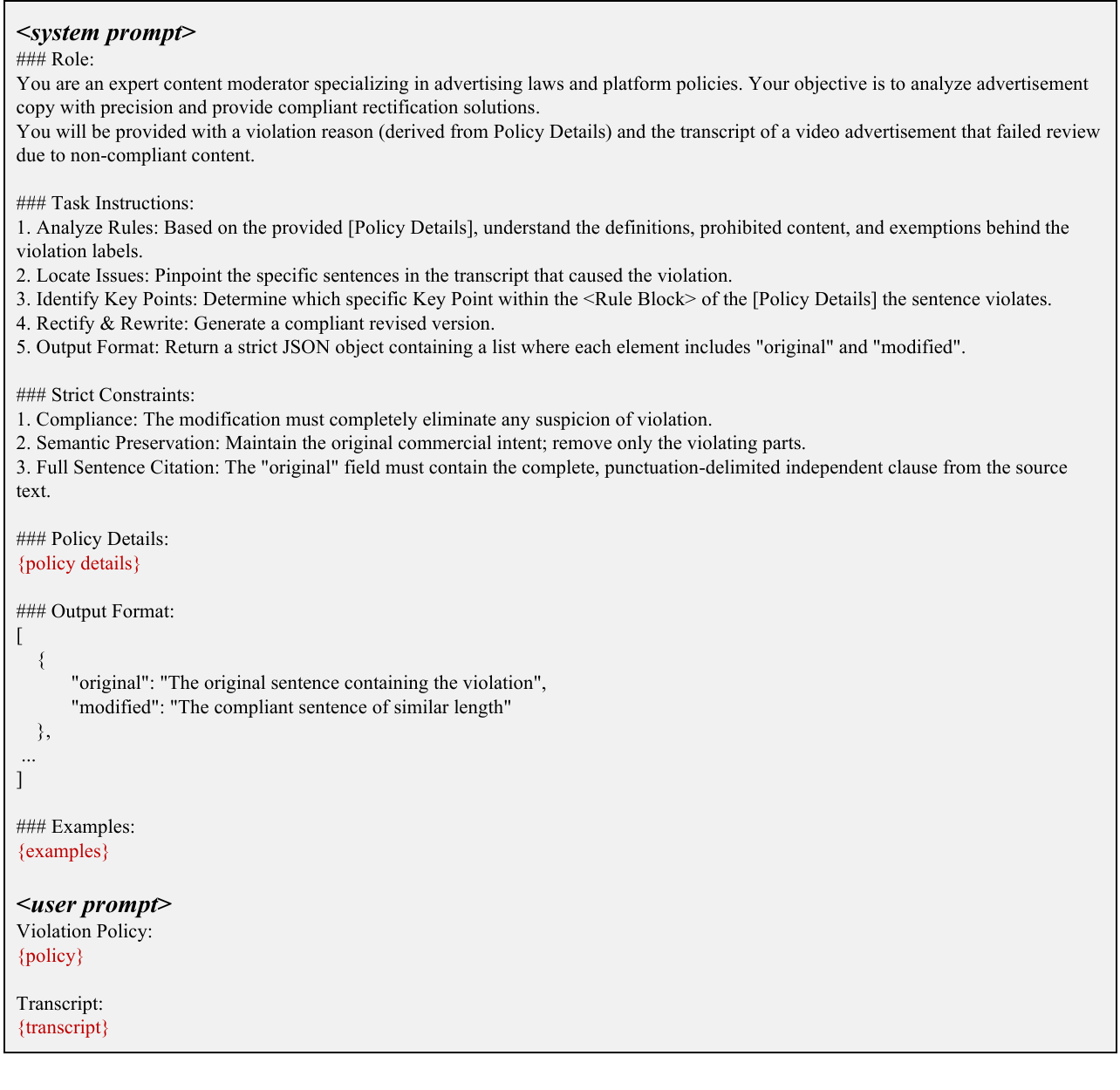} 
  \caption {Standard Prompt for Rectification}
  \label{fig:sft_prompt}
\end{figure*}

Figure \ref{fig:sft_prompt} illustrates the standard prompt template used to guide the LLM in rectifying non-compliant text and generating structured outputs.

\subsection{Prompt for Group-Relative Compliance Experience Extractor}
\begin{figure*}[t]
  
  \includegraphics[width=1\linewidth]{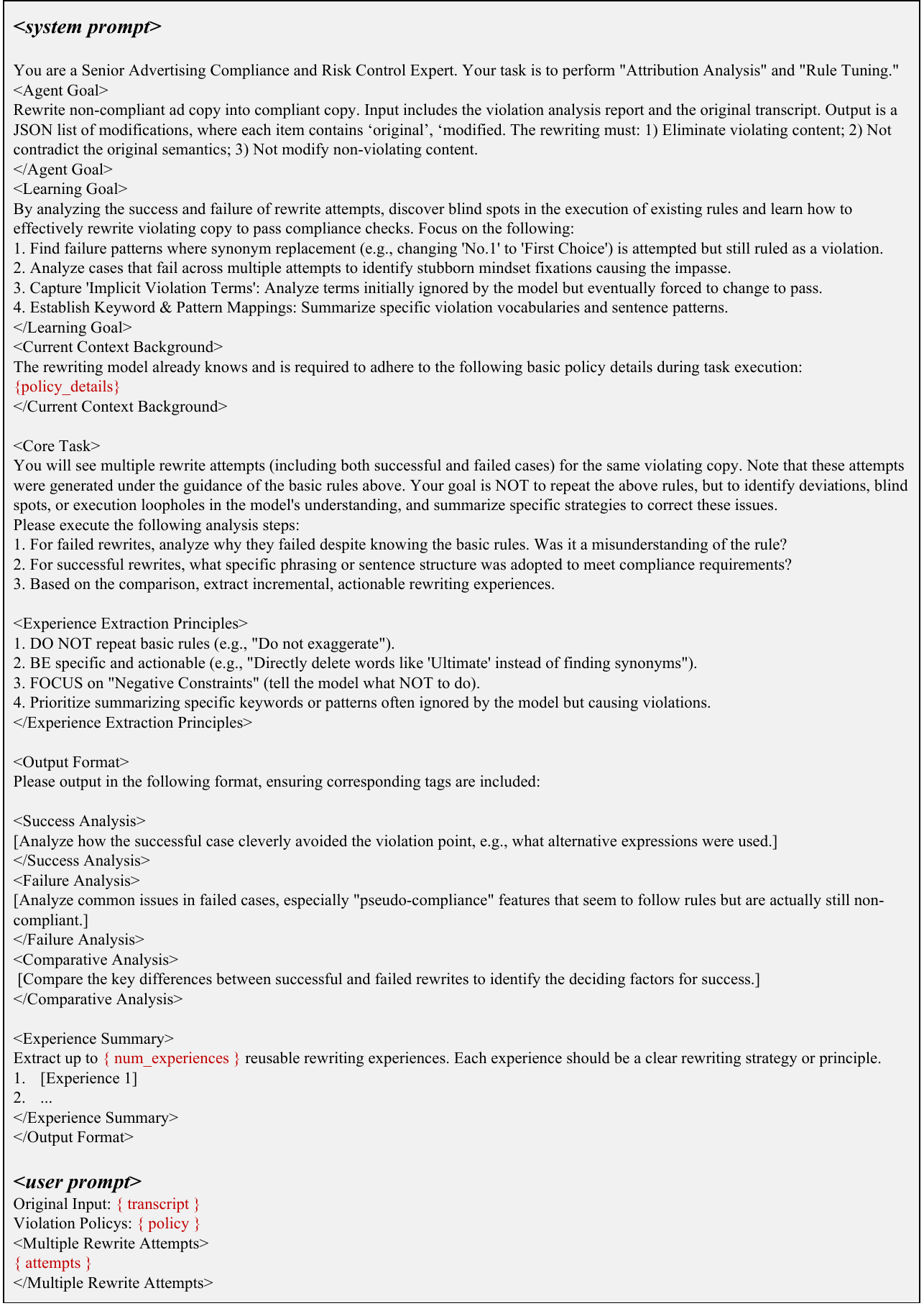} 
  \caption {Prompt for Group-Relative Compliance Experience Extractor}
  \label{fig:gcee_prompt}
\end{figure*}

Figure \ref{fig:gcee_prompt} presents the prompt utilized by the Group-Relative Compliance Experience Extractor.

\subsection{Case Study: Trajectory Splitting and Experience Extraction}
\label{sec:app_example_GCEE}
\begin{figure*}[t]
  
  \includegraphics[width=1\linewidth]{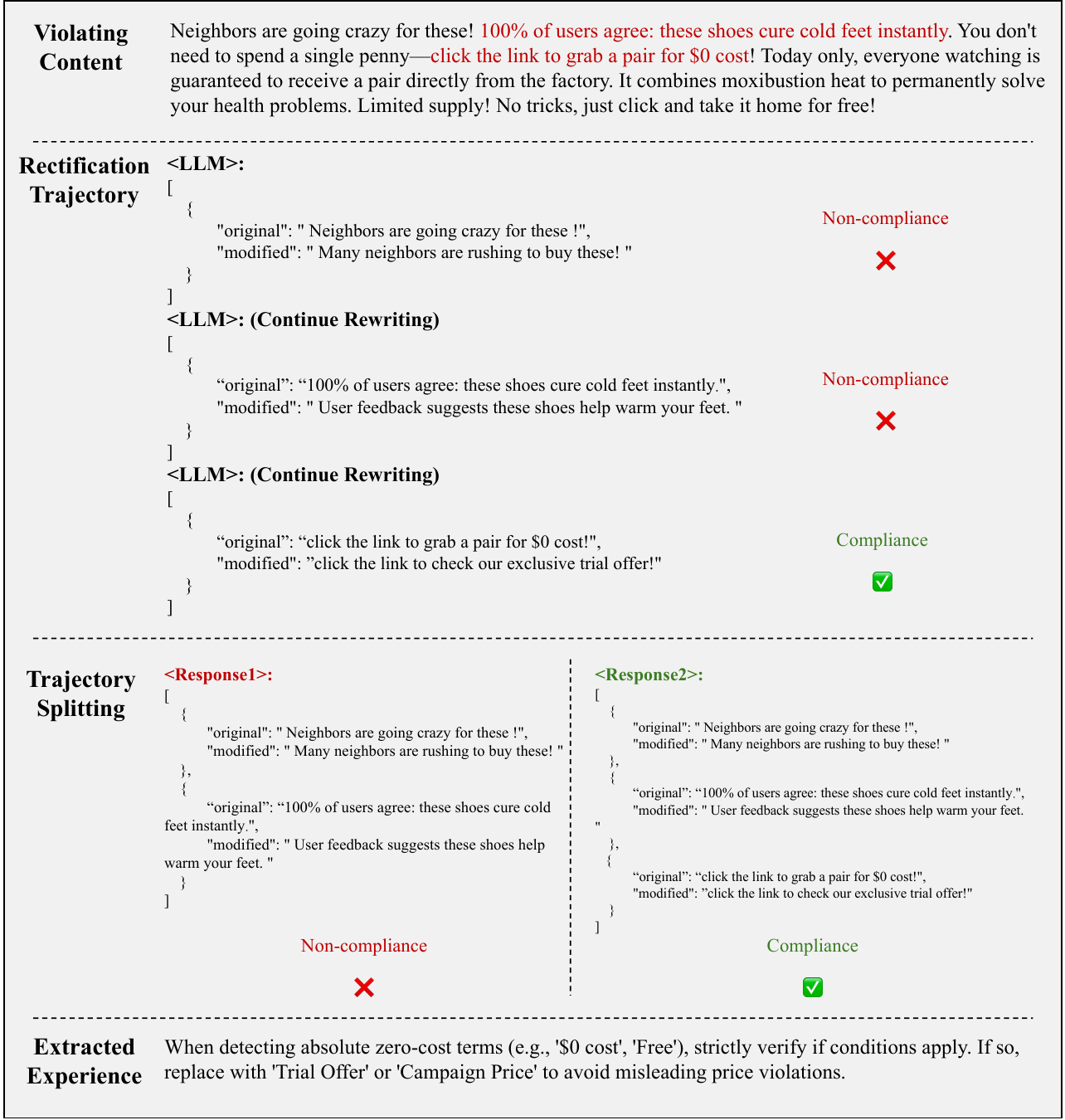} 
  \caption {A complete workflow of trajectory splitting and experience extraction}
  \label{fig:trajectory_case}
\end{figure*}

Figure \ref{fig:trajectory_case} demonstrates a complete workflow of trajectory splitting and experience extraction, where the LLM performs three consecutive rounds of rewriting, incrementally resolving non-compliance based on the previous output. This continuous trajectory is split into contrastive pairs. The extractor then analyzes the semantic gap between these paired responses to extract specific compliance experiences.

\subsection{Algorithmic Details of GCEE}

The concrete hyperparameters for the GCEE data-synthesis stage are as follows: we run 3 epochs with batch size 10; for each sample, we generate 5 rollouts; and we extract at most 2 experiences per sample.

\section{Details of Coherence Evaluation}
To ensure the rectified text maintains linguistic fluency and semantic intent, we employ Gemini3-Flash as an LLM-as-a-judge evaluator. We designed a rigorous three-level evaluation rubric to assess the quality of modifications, mapping each quality level to a specific scalar reward. To align the LLM's judgment with human preferences, we include representative few-shot examples in the prompt to guide the evaluation. Table~\ref{tab:coherence_rubric} details the specific criteria used by the evaluator. 

\input{resources/tables/table_coherence_eval}

\section{Implementation Details}
\label{sec:impl_details}

We build $\mathcal{R}^3$ on top of Qwen3-8B. We perform SFT with LoRA ($r{=}64$) using a learning rate of $1\times10^{-4}$, batch size 64, and train for 2 epochs. We then apply GRPO with the two-phase curriculum; each phase is trained for 1 epoch with a learning rate of $1\times10^{-6}$, group size 8, and a KL coefficient of 0.01. All experiments are conducted on 8 NVIDIA H20 GPUs.

%% file: resources/tables/table_supp_violation_example.tex
\begin{table*}[t]
\centering
\small
\renewcommand{\arraystretch}{1.3}
\begin{tabular}{p{0.18\textwidth} p{0.35\textwidth} p{0.40\textwidth}}
\toprule
\textbf{Policy} & \textbf{Definition} & \textbf{Examples} \\
\midrule
\textbf{Quantitative Claims} & 
Claims that cite specific data, statistics, or probability rates (e.g., success rates, reduction percentages). & 
Our course has successfully helped 5,000 students pass the exam. \\
\midrule
\textbf{Absolute Claims} & 
The use of absolute or exclusive terms that imply a product is indispensable, universal, or has no alternatives. & 
This product is a cosmetic that is essential for every woman. \\
\midrule
\textbf{Superlative Claims} & 
The use of highest-degree adjectives (e.g., "Best," "No.1," "Top") to describe a product's quality or status. & 
We are the No.1 education app in the market. \\
\midrule
\textbf{Misleading \newline Exaggeration} & 
Hyperbolic descriptions that distort the product's actual efficacy. & 
Use this cream and look 20 years younger instantly. \\
\bottomrule
\end{tabular}
\caption{Definitions and examples of the four violation policies targeted in our experiments.}
\label{tab:violation_def}
\end{table*}

%% file: resources/tables/table_coherence_eval.tex
\begin{table*}[h]
\centering
\small
\renewcommand{\arraystretch}{1.5} 
\setlength{\tabcolsep}{8pt}
\resizebox{\textwidth}{!}{
\begin{tabular}{c p{0.42\textwidth} p{0.42\textwidth}}
\toprule
\textbf{Label (Score)} & \textbf{Fluency (Linguistic Quality)} & \textbf{Semantic Intent Preservation} \\ 
\midrule
\textbf{Improvement (1.0)} & 
\textbf{Repairs Flaws.} The edit fixes pre-existing grammatical truncations or logical gaps, resulting in a flow that is significantly more natural than the original. & 
\textbf{Clarifies Intent.} The modification removes ambiguity or awkward phrasing in the original text, making the marketing message clearer and more impactful. \\ 
\midrule
\textbf{Neutral (1.0)} & 
\textbf{Maintains Status Quo.} The sentence structure remains unchanged. Pre-existing typos or colloquialisms are tolerated as long as it is not exacerbated. & 
\textbf{Consistent Intent.} The core marketing message is fully preserved. Modifications are strictly limited to replacing non-compliant terms with compliant synonyms. \\ 
\midrule
\textbf{Degradation (0.0)} & 
\textbf{Introduces Errors.} The edit creates new grammatical faults, awkward collocations, or disjointed connections that did not exist in the source text. & 
\textbf{Distorts Meaning.} The edit reverses the original meaning (e.g., positive to negative), introduces logical contradictions, or loses key product information. \\ 
\bottomrule
\end{tabular}
}
\caption{The evaluation rubric for Coherence Reward. The judge assesses both linguistic fluency and semantic intent preservation to determine the quality of the rectification, assigning a binary-style score}
\label{tab:coherence_rubric}
\end{table*}